\documentclass{article} 
\usepackage{iclr2027_conference,times}

\usepackage{amsmath,amsfonts,bm}

\def\eqref#1{equation~\ref{#1}}

\def\1{\bm{1}}

\DeclareMathAlphabet{\mathsfit}{\encodingdefault}{\sfdefault}{m}{sl}
\SetMathAlphabet{\mathsfit}{bold}{\encodingdefault}{\sfdefault}{bx}{n}

\usepackage{hyperref}
\usepackage{url}
\usepackage{listings}
\usepackage[most]{tcolorbox}
\usepackage{xcolor}
\usepackage{graphicx}
\usepackage[utf8]{inputenc}
\usepackage{textcomp}

\newtcolorbox{flashcard}[1][]{
    enhanced,
    colback=gray!8!white,
    colframe=gray!50!black,
    colbacktitle=gray!30!black,
    coltitle=white,
    fonttitle=\bfseries,
    title=#1,
    rounded corners,
    boxrule=1pt,
    arc=8pt,
    drop shadow,
    width=\linewidth,
    center,
    left=10pt, right=10pt, top=10pt, bottom=10pt
}
\usepackage{booktabs}
\usepackage{array}
\newif\ifshowrevisions
\showrevisionsfalse
\definecolor{revblue}{RGB}{0,70,190}
\ifshowrevisions
    \newcommand{\rev}[1]{\textcolor{revblue}{#1}}
    \newenvironment{revised}{\color{revblue}}{}
\else
    \newcommand{\rev}[1]{#1}
    \newenvironment{revised}{}{}
\fi

\title{
    DuoCook: An Overcooked-Inspired Benchmark for Visual Planning and Coordination
}

\author{\begin{tabular}{@{}p{0.47\textwidth}@{}p{0.47\textwidth}@{}}
Ziyi Jiao & Heng Zhang \\
\mdseries University of British Columbia & \mdseries University of British Columbia \\[2ex]
Drishti Regmi & Xiaokang Ye \\
\mdseries University of California, San Diego & \mdseries University of California, San Diego \\[2ex]
Lianhui Qin & Peter Yichen Chen \\
\mdseries University of California, San Diego & \mdseries University of British Columbia
\end{tabular}}

\iclrfinalcopy
\begin{document}

\maketitle
\lhead{Preprint.}

\begin{abstract}
Coordinating two players in a cooking task requires a vision-language agent to track changing scene states, sequence recipe steps, and allocate shared resources under a time budget. We introduce an Overcooked-inspired benchmark for evaluating these capabilities in Unreal Engine environments constructed using coding agents. A single vision-language model controls both players from top-down visual observations \rev{through separate per-player decision calls}, selecting high-level actions \rev{and image-labeled target cells} while the engine executes navigation. The benchmark contains 20 maps in five categories. These categories support evaluation of \rev{two-player} coordination under equipment contention, changing terrain, and handoff dependencies. A prop legend and pot ingredient-count indicators provide explicit visual cues, while grid-based destinations support spatial grounding\rev{; text gives no station coordinates, so target locations come mainly from the image}. The evaluation compares \rev{five VLMs and a privileged scripted reference policy on all 20 maps}, measuring task score, order completion, delivery precision, \rev{aim accuracy}, and model API expenditure. \rev{Two paired ablations on all 20 maps toggle a shared self-authored notebook and reduce the textual prompt from about 3,600 to about 440 words while keeping the images fixed.} \rev{The strongest model completes 35\% of orders and exceeds the scripted reference, which has ground-truth station access; the notebook's effect is mixed in sign and not statistically reliable, and reducing the prompt text lowers the score of two models while leaving the other three nearly unchanged.} Together, the environment design and evaluation protocol provide a framework for investigating how visual state tracking and task scheduling contribute to coordinated cooking behavior.
\end{abstract}

\section{Introduction}
\label{sec:introduction}

A cooking task can fail even when individual actions appear sensible. A player may chop the correct ingredient while blocking a board needed for an urgent order, or wait beside a boiling pot instead of working elsewhere. Efficient cooking requires tracking unfinished operations, assigning complementary work, and revising schedules as equipment and ingredients become available. This raises a broader question: can a vision-language model (VLM) use visual observations and structured task information to coordinate interdependent work under order deadlines?

Overcooked-AI examines cooperation with humans \citep{carroll2019utility}, MindAgent studies multi-agent task allocation \citep{gong2024mindagent}, and Collab-Overcooked introduces resource isolation and process-oriented evaluation \citep{sun2025collab}. Visual-agent benchmarks evaluate interaction across games, embodied environments, and graphical interfaces \citep{paglieri2025balrog, liu2025visualagentbench, yang2025embodiedbench}. We study \rev{two-player} scheduling across kitchens with different spatial and operational constraints, using a common multimodal interface to connect kitchen state, recipes, orders, and player assignments.

We introduce DuoCook, an Overcooked-inspired benchmark in Unreal Engine where a single VLM controls two players. Its 20 maps span five categories: basic, shared equipment, dynamic, dynamic with shared equipment, and separated workspaces (Section~\ref{sec:maps}), introducing equipment contention, terrain changes, and required handoffs. The model \rev{chooses each player's next action in a separate decision call} to maximize \rev{a shared} delivery score under deadlines, time-dependent tips, and a streak bonus, making \emph{\rev{two-player} coordination} the evaluation target. Coding agents construct kitchens through SimWorld Studio \citep{kang2026simworld}, followed by manual refinement and actor-contract verification (Section~\ref{sec:generation}).

\rev{Each call receives} a top-down kitchen image \rev{with grid labels}, prop legend, task instructions, live-state information \rev{including the partner's current operation}, and \rev{the full operation vocabulary}. \rev{Text gives no station coordinates, pot contents, or terrain configuration, so target cells come mainly from the image.} The model selects actions and destination coordinates while the engine handles navigation. State changes trigger model calls, with simulation paused during model inference. The interface thus evaluates \rev{image-grounded} destination selection and scheduling with structured state assistance\rev{, not low-level control or real-time reaction}.

The evaluation protocol compares \rev{five VLMs and a privileged scripted reference policy on all 20 maps}, reporting total score, order completion rate, delivery precision, \rev{aim accuracy}, and API expenditure per episode \rev{(320 episodes)}. \rev{Two paired ablations on all 20 maps toggle a shared, self-authored notebook and replace the full textual prompt with a minimal one while keeping the images fixed. The strongest model, \texttt{gpt-6-astra}, completes $35.0\%$ of orders and delivers on 18 of 20 maps, above the scripted reference ($21.3\%$) that reads ground-truth station locations; every map receives at least one correct delivery from some model. The notebook's effect differs in sign across models and is not statistically reliable, and the minimal prompt lowers the score of two of five models.}

The paper makes two contributions:
\begin{itemize}
    \item A benchmark of 20 agent-generated kitchens for \rev{single-model two-player} planning and coordination, with five map categories and a shared multimodal interface\rev{, together with construction records for the generated maps}.
    \item An evaluation protocol combining \rev{performance, command-validity}, and API-cost metrics with \rev{a privileged scripted reference and paired ablations of shared memory and prompt text}.
\end{itemize}

\section{Related Work}
\label{sec:related}

\subsection{Cooking Environments and Cooperative Agents}

Overcooked-AI studies coordination and cooperation with humans \citep{carroll2019utility}, while ProAgent uses language models to infer teammate intentions \citep{zhang2024proagent}. MindAgent's CuisineWorld supports centralized multi-agent scheduling \citep{gong2024mindagent}, and Collab-Overcooked combines resource isolation, asymmetric knowledge, and communication to evaluate interdependent collaboration \citep{sun2025collab}. DuoCook builds on these precedents with one visually grounded model controlling both players and map variations that expose resource and spatial dependencies. It evaluates \rev{coordination between two players driven by the same model under a shared score}, rather than inference about an independently controlled partner; neither multi-cook control nor required handoffs alone constitutes its contribution. \rev{What DuoCook adds is their combination in one rendered 3D interface: targets are localized mainly from the image, because text gives no station coordinates, and the suite includes layouts with equipment contention, moving terrain, and mandatory handoffs. A privileged scripted reference (Section~\ref{sec:scripted}) shows that the workloads are attainable. Appendix~\ref{app:comparison} compares these properties with related environments.}

ParaCook~\citep{zhang2026paracook} represents cooking tasks as directed acyclic graphs and measures concurrent execution efficiency. Robotouille~\citep{gonzalez2025robotouille} examines asynchronous, long-horizon planning, while CookBench~\citep{cai2025cookbench} combines recipe preparation with customer-intention recognition. Related planning benchmarks include WorfBench~\citep{qiao2025benchmarking} and Plan-over-Graph~\citep{zhang2025plan} for graph-structured workflows, AsyncHow~\citep{3692070.3693283} for ordering tasks with known durations, and TimeArena~\citep{zhang2024timearena} for scheduling in an interactive environment. PACE~\citep{huang2026pace} overlaps planning and execution through interleaved reasoning and action. These directions inform DuoCook's scheduling demands, such as preparing another dish while soup boils.

\subsection{Visual Interaction and Game Benchmarks}

Balrog evaluates planning and exploration across games \citep{paglieri2025balrog}, while VisualAgentBench spans embodied interaction, graphical interfaces, and visual design \citep{liu2025visualagentbench}. EmbodiedBench covers semantic tasks, navigation, and manipulation \citep{yang2025embodiedbench}; VideoGameBench evaluates raw visual gameplay, including a paused setting that separates some inference-latency effects from action selection \citep{zhang2025videogamebench}. DuoCook focuses on matched interventions within cooking workflows. Its engine-executed navigation, grid labels, and structured state assist the controller, so scores are not directly interchangeable with those from different observation and control interfaces.

\subsection{Persistent Context for Interactive Agents}

ReAct interleaves reasoning and action using interaction feedback \citep{yao2023react}, while Reflexion stores verbal reflections across trials \citep{shinn2023reflexion}. \rev{In DuoCook, each player always receives its own six most recent decisions with their target cells and acceptance feedback. The memory ablation adds a shared, decaying notebook written by the players themselves (Section~\ref{sec:memory}), and tests its effect on performance and inference cost.}

\subsection{Generated Environments for Evaluation}

Generated environments can support agent evaluation or serve as the output of a generative model. Procgen uses procedurally generated levels to evaluate agent generalization \citep{cobbe2020leveraging}, while MarioGPT generates Mario levels from text \citep{sudhakaran2023mariogpt}. In DuoCook, coding agents build Unreal Engine kitchens from reference layouts and construction instructions, followed by manual refinement and actor-contract verification (Section~\ref{sec:generation}). This process supplies the environments for evaluating cooking controllers.

\section{System and Task Design}
\label{sec:system}

\begin{figure}[t]
    \centering
    \includegraphics[width=\linewidth]{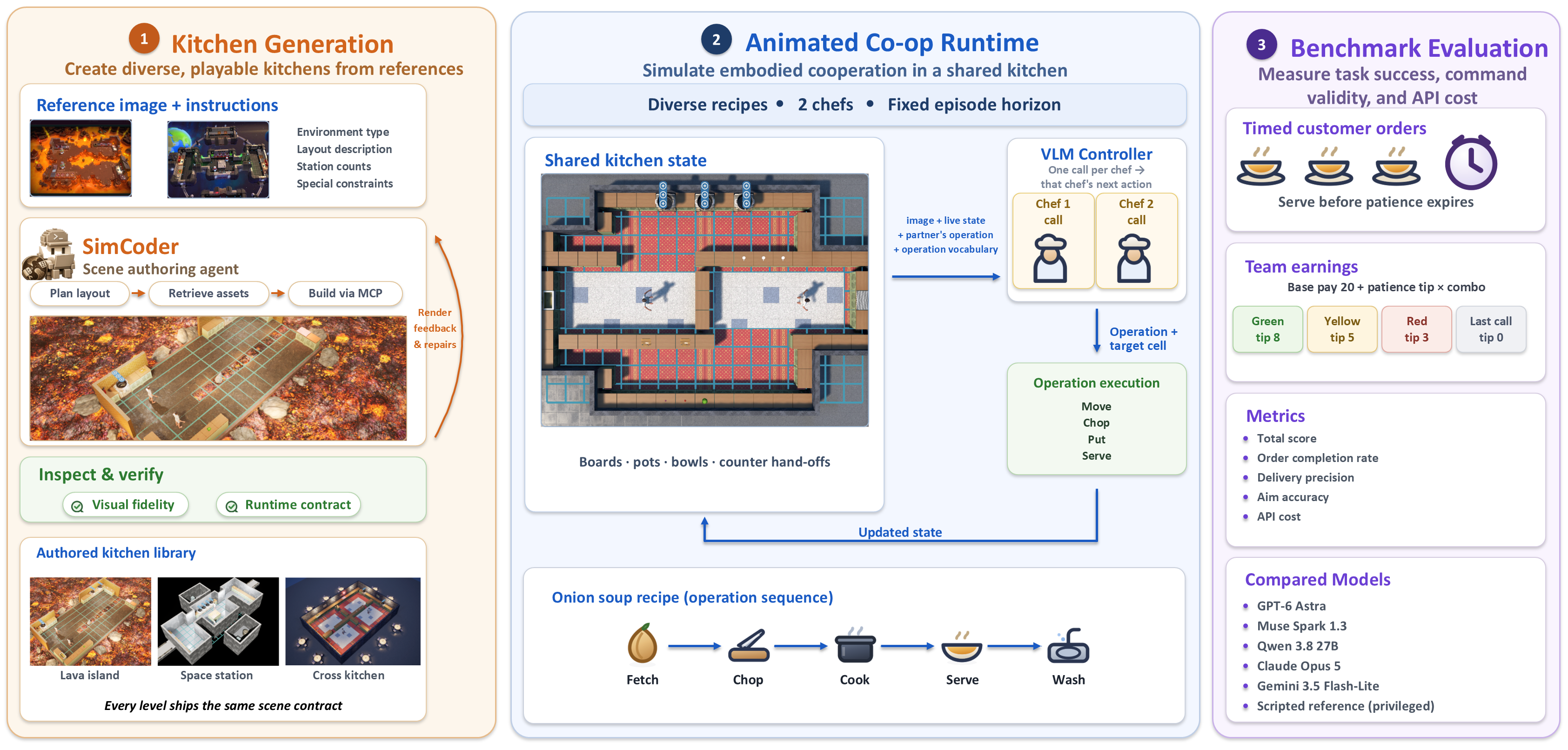}
    \caption{Benchmark pipeline. Left: coding-agent kitchen construction, manual refinement, and actor-contract verification. Center: \rev{one VLM controlling two players through per-player decision calls on shared multimodal inputs}, illustrated by onion-soup preparation. Right: timed-order scoring and evaluation of score, order completion, delivery precision, \rev{aim accuracy}, and API expenditure.}
    \label{fig:pipeline}
\end{figure}

Figure~\ref{fig:pipeline} summarizes the benchmark workflow, from kitchen construction to controller execution and evaluation.

\subsection{Map Generation with Coding Agents}
\label{sec:generation}

We construct kitchens in Unreal Engine using SimWorld Studio, which provides coding agents with Model Context Protocol (MCP) tools and Markdown skills for scene configuration and asset placement. Two separate \texttt{claude-opus-5} processes author the generation prompt and build the scene, respectively. The authoring process reviews the SimWorld Studio architecture, then receives the game context, a reference image, and an actor naming and scene-structure contract. It produces a construction prompt with a completion checklist covering required appliances and VLM-controlled MetaHuman characters. The construction process executes this prompt with access to the game context, pipeline code, actor contract, and reference image. We then refine the scene manually and run a script to verify required actors and their names. Manual edits address decoration, collisions, visibility, and camera configuration. In a truck scene, for example, we moved rocks clear of the truck bed, adjusted the top-down camera's yaw, added an isometric camera, and removed a truck wall to expose the kitchen.

\begin{revised}
\paragraph{Construction records.} The construction agent edits a live editor through an MCP tool that executes short Python scripts, and every call is logged. Across the surviving logs for 27 constructed levels, which include levels outside the benchmark, construction took 108 agent runs, \$352.70 of model cost, and 25.3 agent-hours. Per level, this was a median of 3 runs, \$11.65, 81 script calls, and 1.2 hours of active operator time. These figures are cumulative and include discarded builds: 25 of the 27 levels had their prompt submitted more than once, and 8 of the 19 benchmark levels with records needed two or more correction episodes. Logs do not distinguish operator fixes applied through the same tool from agent edits, so we report operator time rather than an autonomous success rate. Appendix~\ref{app:generation} gives per-level records and recurring failure modes.
\end{revised}



\subsection{Map Conditions}
\label{sec:maps}

The benchmark contains 20 functional maps based on Overcooked layouts, grouped by equipment sharing, terrain changes, and item transfers\rev{: five basic, four shared-equipment, four dynamic, four dynamic with shared equipment, and three separated-workspace maps. All experiments use all 20 maps.} Appendix~\ref{app:maps} lists source levels \rev{and categories}.


All maps contain floors, counters, ingredient bins, pots, plates, chopping boards, and serving points. Bins, boards, and pots are fixed to counters, which block movement but support food and plate placement. Players move, carry and place objects, chop ingredients, plate food, and serve dishes. After serving, dirty plates appear in sinks for washing on sink-equipped maps; otherwise, clean plates reappear at their initial locations. The map categories vary the coordination required by these operations. Basic and shared-equipment maps have fixed terrain: the former allow independent order completion, while the latter require players to share stations such as chopping boards or pots. Dynamic maps introduce counters or floor sections that move back and forth every \rev{500} steps, changing obstacles or traversable terrain; gaps remain inaccessible, preventing falls. Four dynamic maps require equipment sharing, and four do not. In separated workspaces, players cannot cross between regions and must pass ingredients or plates across counters, so neither can complete a dish alone.


\subsection{Cooking Task and Controller Objective}
\label{sec:task}

A single VLM controls two MetaHuman players to maximize the score from preparing and serving orders within the episode horizon (Section~\ref{sec:setup}). Menus include salads and soups, with recipes and order queues configured by level. Chopping and dish washing occupy a player throughout execution, whereas soup boils unattended, allowing concurrent work.

\paragraph{Orders and scoring.}
Orders arrive every 1,000 steps\rev{, starting at step 0, so each 4,000-step episode seats four orders, the last with 1,000 steps remaining. Each map's menu is cycled in a fixed order.} Orders expire after 3,000 steps and may be fulfilled in any order. \rev{A delivered dish goes to the longest-waiting order for that dish; a dish that matches no waiting order earns nothing.} Each delivery earns 20 points plus a time-dependent tip scaled by a combo multiplier:
\begin{equation}
    R_i = 20 + m_i b(w_i),
    \label{eq:order_reward}
\end{equation}
where $w_i$ is the waiting time at delivery, $b(w_i)$ is the base tip (Table~\ref{tab:order_tips}), and $m_i$ follows the schedule $(1,1,2,3,4)$, capped at $4$. Deliveries build the combo streak; an expired order resets the streak to zero. The multiplier applies only to tips.

\begin{table}[t]
    \centering
    \caption{Base tips by elapsed waiting time for a non-expired order. The thresholds correspond to 45\%, 70\%, and 85\% of the 3,000-step waiting limit.}
    \label{tab:order_tips}
    \begin{tabular}{llr}
        \hline
        Zone & Elapsed waiting time $w$ (steps) & Base tip \\
        \hline
        Green & $0 \leq w \leq 1{,}350$ & 8 \\
        Yellow & $1{,}350 < w \leq 2{,}100$ & 5 \\
        Red & $2{,}100 < w \leq 2{,}550$ & 3 \\
        Last call & $2{,}550 < w < 3{,}000$ & 0 \\
        \hline
    \end{tabular}
\end{table}



\subsection{Multimodal Agent Inputs}
\label{sec:observation}

\begin{figure}[t]
    \centering
    \includegraphics[width=\linewidth]{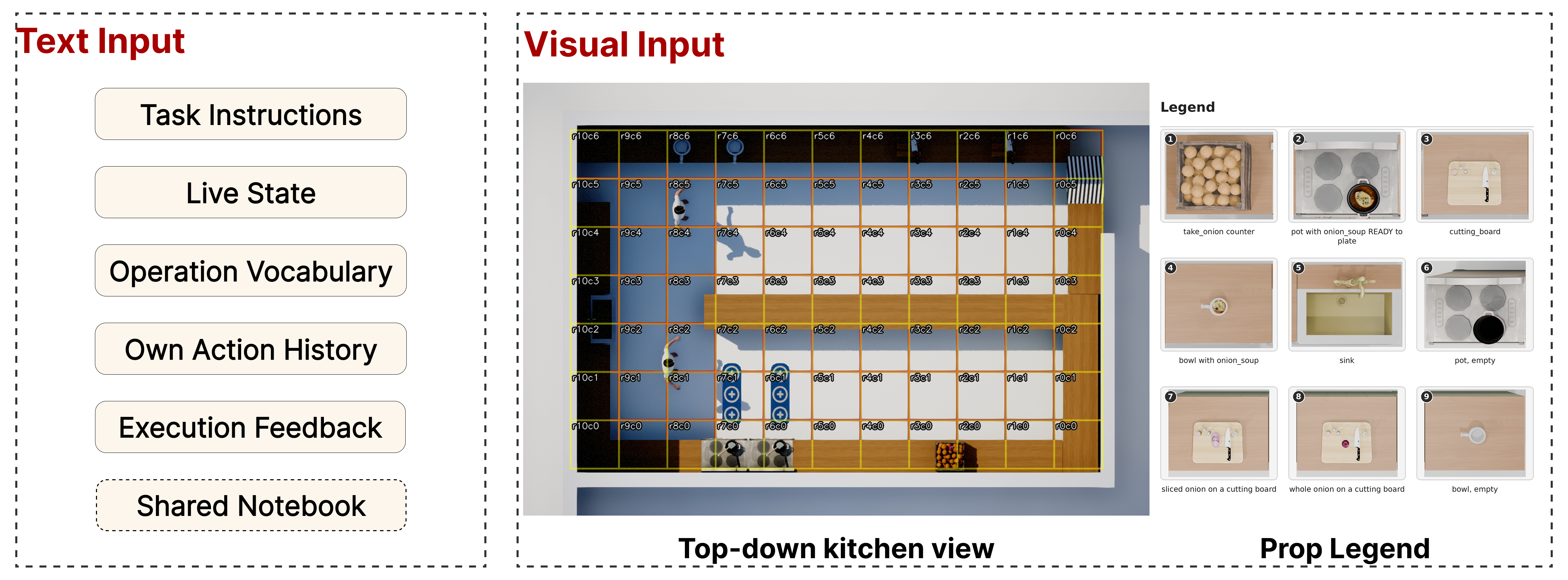}
    \caption{\rev{Inputs to one player's decision call.} Left: instructions, live-state JSON, \rev{the operation vocabulary, the player's private six-decision history, the shared notebook in the memory condition,} and execution feedback. Right: a top-down kitchen image with grid labels and pot indicators, plus a prop legend. \rev{Table~\ref{tab:channels} lists the information carried by each channel.}}
    \label{fig:agent_inputs}
\end{figure}

The controller receives kitchen images alongside task instructions, live state, \rev{the operation vocabulary}, execution feedback, and \rev{its own} action history (Figure~\ref{fig:agent_inputs}). \rev{The same VLM is queried separately for each player whenever that player's current operation finishes, fails, or stalls. Each call is stateless: anything carried across decisions is supplied explicitly as text (Table~\ref{tab:channels}).}

\rev{Each call contains} a top-down kitchen image with grid labels \rev{(about 90 pixels per cell)} and a prop legend. A map-specific system prompt of approximately \rev{3,600 words (4,000 with the notebook)} provides kitchen rules and recipes, using the simulator's menu, and is re-sent at every decision. \rev{The live-state JSON reports waiting orders with remaining time and tip zone, earnings, combo streak, the player's own cell and held item, and the partner's current operation, target cell, and claimed stations. It contains no station or object coordinates, no pot or board contents, and no partner position or held item.} Game-UI information is supplied as text to reduce dependence on UI appearance. \rev{The prompt lists the full operation vocabulary at every call without filtering by feasibility, so the model must decide which operations are possible and where, mainly from the image.} Rejection feedback explains failures without suggesting corrective plans \rev{and never reveals the correct cell}. \rev{Each player also receives its own six most recent decisions, newest first, with target cells and execution acceptance; this history is private to the player and present in all conditions.}

\begin{table}[t]
    \centering
    \small
    \caption{\rev{Information channels in each decision call under the full prompt, from the benchmark code; the minimal prompt removes the partner fields. ``Image only'': not stated in text in the current call. No text channel lists station coordinates, but accepted targets in the history, the partner's target cell, and notebook rows carry some spatial information.}}
    \label{tab:channels}
    \rev{\begin{tabular}{p{0.56\linewidth}p{0.36\linewidth}}
        \toprule
        Information & Channel \\
        \midrule
        Station and object locations; which cells hold ingredients, plates, or dishes & Image only \\
        Pot fill and readiness; board contents; moved terrain; partner position and held item & Image only \\
        Orders, remaining time, tip zone, earnings, combo streak & Live-state JSON \\
        Own cell and held item; partner's current operation, target cell, and claims & Live-state JSON \\
        Recipes, rules, scoring; full operation vocabulary (unfiltered) & System prompt \\
        Own last six decisions with acceptance; failure reason & Text, private to the player \\
        Shared notebook (memory condition only) & Text, shared by both players \\
        \bottomrule
    \end{tabular}}
\end{table}

\subsection{Grid-Based Navigation and High-Level Actions}
\label{sec:navigation}

Each map has an $N\times M$ grid with coordinates \texttt{rxcy}, where $x$ and $y$ denote row and column. The origin is randomly assigned per map and then fixed; the controller reads image labels to locate targets (Figure~\ref{fig:agent_inputs}). Commands pair a destination coordinate with an action, such as pickup, chopping, or serving. Station interactions target the station cell; the engine handles the approach using a navigation mesh (NavMesh) that excludes counters and gaps. Incompatible action-target pairs and unreachable destinations return errors for the controller to revise and retry. \rev{A target within one orthogonal cell of the correct station is accepted, which the model is not told. A player's decision call is triggered when its current operation finishes, fails, or stalls.} The engine updates observations and pauses simulation during VLM inference. When one player finishes, the controller assigns its next action while preserving the other player's ongoing action, which resumes after inference. \rev{When moving terrain blocks a route, the engine replans; a walk that cannot complete is reported to the model as a failure, and the model then selects a new destination and action from the updated scene.}



\section{Experiments}
\label{sec:experiments}

Our evaluation examines how a single model coordinates two players under equipment-sharing constraints, changing terrain, and required item transfers. It consists of the main benchmark\rev{, a privileged scripted reference, and two paired ablations, all on the complete 20-map suite}.

\subsection{Evaluation Setup}
\label{sec:setup}

The complete benchmark contains 20 functional maps\rev{, and every model, policy, and condition is evaluated on all of them} (Appendix~\ref{app:maps}). \rev{Score, completion, aim accuracy, failures, and cost are equal-weight means over the 20 maps; precision pools episodes.}

\paragraph{Evaluated models.}
The model roster consists of \texttt{gpt-6-astra}, \texttt{claude-opus-5}, \rev{\texttt{muse-spark-1.3}}, \texttt{qwen3.8-27b}, and \texttt{gemini-3.5-flash-lite}. \rev{The same model controls both players through per-player decision calls (Section~\ref{sec:observation})}, and the two players can act concurrently. \rev{\texttt{qwen3.8-27b} is accessed through OpenRouter, and the other four models through an OpenAI-compatible API gateway. All runs enable reasoning at low effort.}


\paragraph{Episode protocol.}
Each episode runs for 4,000 steps\rev{, where a step is one simulator tick}. Controller calls follow state-change events, and the simulation pauses during inference, as described in Section~\ref{sec:navigation}.

For each map, the initial state, number of orders, and order sequence are fixed and shared across models and \rev{all} conditions. \rev{Each model runs once per map in each of three conditions: full prompt with the shared notebook (the main condition), full prompt without the notebook, and minimal prompt without the notebook. Together with the scripted reference, this gives 320 episodes.} These settings match the starting conditions and order workload; subsequent trajectories depend on the model's actions. For \rev{each} ablation, we compute the difference between conditions on each of the \rev{20} maps and then average these paired differences. With one run per model, map, and condition, results are descriptive and do not estimate variability across repeated model runs\rev{; $\pm$ values are standard errors across maps, and we test paired effects with an exact two-sided sign test over maps}.


\subsection{Evaluation Metrics}
\label{sec:metrics}

We use five metrics to measure task performance, \rev{command validity}, and inference cost. The same metrics apply to \rev{all conditions}.

\paragraph{Total score.}
The episode score measures accumulated points under the scoring rules in Section~\ref{sec:task}. Each fulfilled order earns the base reward and combo-adjusted tip in Equation~\ref{eq:order_reward}; orders may be fulfilled in any sequence.

\paragraph{Order completion rate.}
Let $N_{\mathrm{completed}}$ denote the number of completed orders and $N_{\mathrm{orders}}$ the total number of orders in an episode. Order completion rate is
\begin{equation}
    C = \frac{N_{\mathrm{completed}}}{N_{\mathrm{orders}}}.
    \label{eq:outcomes}
\end{equation}
This metric measures how much of the order workload the controller completes within the episode horizon. \rev{Every episode seats $N_{\mathrm{orders}}=4$ orders, including the last order, which arrives with 1,000 steps remaining.}

\paragraph{Delivery precision.}
Let $N_{\mathrm{correct}}$ denote the number of correct deliveries and $N_{\mathrm{served}}$ the total number of dishes served. Delivery precision is
\begin{equation}
    P = \frac{N_{\mathrm{correct}}}{N_{\mathrm{served}}}.
\end{equation}
\rev{A delivery is correct if it matches a waiting order. Because most episodes serve few or no dishes, we pool $P$ over all episodes of a controller and report the underlying counts. $P$ characterizes only serving attempts and does not identify why orders remain unfinished.}

\paragraph{\rev{Aim accuracy.}}
\rev{Each model command names an operation and a target cell read from the image. A command is graded valid if the operation is possible and the target lies on, or within one orthogonal cell of, the correct station. Commands with an impossible operation, a missing cell, or a wrong cell are graded invalid. Aim accuracy is the fraction of graded commands that are valid, averaged over episodes. It reflects target selection, which relies mainly on the image, together with operation feasibility, and cannot isolate image misreading; no model episode has zero graded commands. We also report failed actions per episode, which additionally include navigation failures and malformed responses.}

\paragraph{Model execution cost.}
Model execution cost is the total amount spent on model API calls during an episode, reported in US dollars (USD). It includes all billed calls, including retries, and any charges for image inputs and action-history context. \rev{The API gateway does not report billing, so we estimate it from recorded tokens at list prices (Appendix~\ref{app:cost}).} We report mean API expenditure per episode alongside task performance.

\subsection{\rev{Privileged Scripted Reference}}
\label{sec:scripted}

\rev{To calibrate difficulty, we run a fixed-priority scripted policy under the same horizon, initial states, order schedules, and execution engine. It is \emph{privileged}: it reads the simulator's feasible operations and ground-truth station locations directly, so it never aims. When holding an item, it prefers to deliver, wash, plate, fill a bowl or pot, or place on a board, in that order. With empty hands, it takes a finished dish, a dirty bowl, a clean bowl when a pot is ready, a sliced or whole ingredient, or cuts. Ties go to the oldest order. The two players prefer different station indices and skip stations claimed by the partner, and counter transfers have the lowest priority. The policy is not tuned per map. It is a feasibility reference with perfect perception and greedy scheduling, not an upper bound.}

\subsection{Main Benchmark Evaluation}
\label{sec:mainresults}

\begin{table}[t]
    \centering
    \small
    \caption{\rev{Main condition (full prompt, shared notebook), equal-weight means over all 20 maps, one episode per map; $\pm$ is the standard error across maps. Compl.: order completion (\%). Prec.: delivery precision (\%), pooled over episodes as correct/served dishes. Maps: maps with at least one correct delivery. Aim: aim accuracy (\%). Fail: failed actions per episode. Cost: estimated API cost (USD) per episode. The scripted reference has privileged state and no notebook.}}
    \label{tab:main}
    \setlength{\tabcolsep}{4pt}
    \rev{\resizebox{\linewidth}{!}{\begin{tabular}{lrrrrrrr}
        \toprule
        Controller & Score $\uparrow$ & Compl.\ $\uparrow$ & Prec.\ $\uparrow$ & Maps $\uparrow$ & Aim $\uparrow$ & Fail $\downarrow$ & Cost $\downarrow$ \\
        \midrule
        \texttt{gpt-6-astra} & \textbf{35.1}$\pm$5.0 & \textbf{35.0} & 100 (28/28) & \textbf{18} & \textbf{84.7} & \textbf{19.3} & 5.41 \\
        \texttt{muse-spark-1.3} & 28.6$\pm$5.2 & 28.8 & 100 (23/23) & 16 & 75.6 & 27.0 & 1.09 \\
        \texttt{qwen3.8-27b} & 12.8$\pm$3.9 & 12.5 & 91 (10/11) & 8 & 55.1 & 46.8 & \textbf{0.30} \\
        \texttt{claude-opus-5} & 8.2$\pm$3.4 & 8.8 & 100 (7/7) & 5 & 70.6 & 36.7 & 4.61 \\
        \texttt{gemini-3.5-flash-lite} & 7.8$\pm$2.7 & 7.5 & 50 (6/12) & 6 & 49.7 & 53.0 & 0.33 \\
        \midrule
        Scripted (privileged) & 20.2$\pm$3.5 & 21.3 & 71 (17/24) & 14 & --- & 30.5 & --- \\
        \bottomrule
    \end{tabular}}}
\end{table}

\begin{figure}[t]
    \centering
    \includegraphics[width=\linewidth]{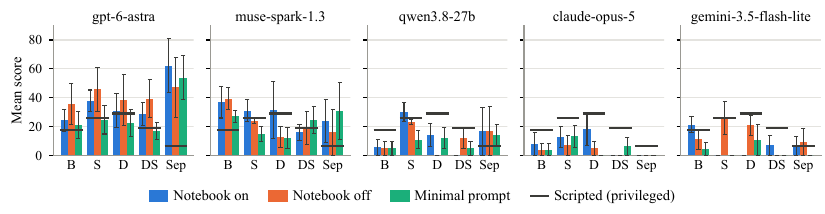}
    \caption{\rev{Mean score by map category for each model and condition (error bars: standard error of the category mean). B: basic, S: shared equipment, D: dynamic, DS: dynamic with shared equipment, Sep: separated workspaces. Black lines mark the privileged scripted reference. Per-map values are in Appendix~\ref{app:permap}.}}
    \label{fig:category}
\end{figure}

\rev{Table~\ref{tab:main} reports the main condition.} The benchmark is far from saturated. \rev{The strongest model, \texttt{gpt-6-astra}, completes $35.0\%$ of orders, followed by \texttt{muse-spark-1.3} at $28.8\%$; the other three complete $7.5$--$12.5\%$. Of the 100 model episodes in this condition, 47 end with no correct delivery, and no episode in any condition completes all four orders. The best observed episode completes three of four orders, on five maps. The map suite nevertheless separates the controllers: \texttt{gpt-6-astra} scores on 18 of 20 maps and \texttt{claude-opus-5} on 5.} Each cell is a single episode, so we report descriptive differences without estimating run-to-run variability.

\begin{revised}
\paragraph{Scripted reference.} The privileged script scores $20.2$, with $21.3\%$ completion, and delivers on 14 maps. It establishes that orders are attainable under the exact horizon and schedule on those maps. On the remaining six (1-1, Kevin 1, 4-1, 4-3, 2-4, and 3-2), at least one model delivers, so every map has at least one demonstrated delivery. Two models exceed the script without its access to ground-truth stations and feasible operations: \texttt{gpt-6-astra} scores higher on 13 maps, lower on 4, and ties on 3. The other three score below it. The script also differs in action priorities and needs no aiming, so the gap cannot be attributed to scheduling alone. The script also serves 7 dishes that no waiting order requested. It fails most often on separated workspaces ($8\%$ completion), consistent with counter transfers having its lowest priority, whereas \texttt{gpt-6-astra} reaches $58\%$ there.

\paragraph{Categories.} No category is unsolved (Figure~\ref{fig:category}). Category difficulty depends on the model. Dynamic maps with shared equipment give the lowest or tied-lowest completion for \texttt{claude-opus-5}, \texttt{muse-spark-1.3}, and \texttt{qwen3.8-27b} (0--19\%). Map 4-3 is the hardest single map, with one correct delivery in 15 model episodes. These category means rest on three to five maps each and are descriptive.

\paragraph{Command validity and task success.} Aim accuracy ranges from $49.7\%$ to $84.7\%$, so between 15\% and 50\% of graded commands target a wrong cell or an impossible operation. It tracks task success only loosely: \texttt{claude-opus-5} has the third-highest aim accuracy ($70.6\%$) but the second-lowest score. Precision is $100\%$ for three models but only $50\%$ for \texttt{gemini-3.5-flash-lite}, which serves dishes that no waiting order requested. \end{revised}Cost spans more than an order of magnitude and does not track performance: \rev{\texttt{muse-spark-1.3} scores second at \$1.09 per episode, a fifth of \texttt{gpt-6-astra}'s \$5.41.}\begin{revised}
\end{revised}


\subsection{Memory Ablation}
\label{sec:memory}

\begin{figure}[t]
    \centering
    \includegraphics[width=\linewidth]{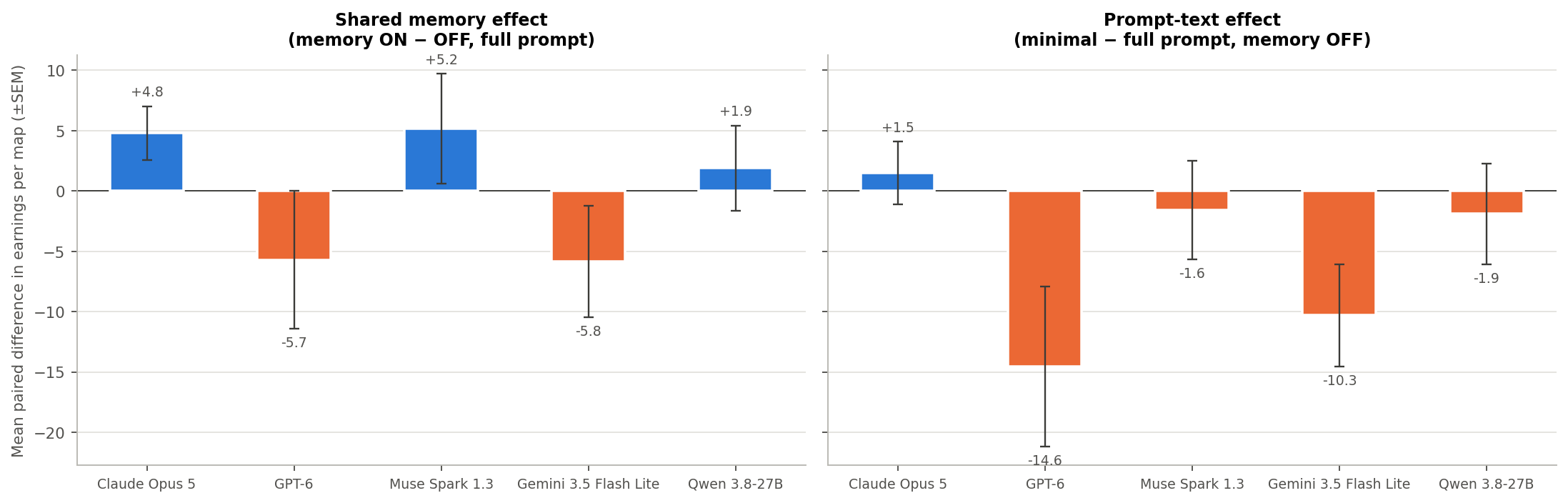}
    \caption{\rev{Paired per-map score differences on all 20 maps. Left: shared notebook on minus off (full prompt). Right: minimal minus full prompt (notebook off). Error bars are standard errors of the mean paired difference across maps; with one episode per cell, they do not capture run-to-run variability.}}
    \label{fig:ablation}
\end{figure}

We compare controllers with and without \emph{shared memory} on \rev{all 20 maps}. The memory-enabled results are reused from the corresponding main-benchmark episodes. For each model and map, the memory-disabled run uses the same initial state and order schedule, with only the shared memory removed.

The shared memory is a notebook the two players fill in themselves. With any decision a player may attach up to three observations, each a tuple of grid cell, object kind, status, and an optional free-text note. Both players read the resulting table back on every subsequent call, and each row carries the identity of whoever wrote it. Entries decay on an exponential forgetting curve with a 600-step half-life and are displayed in three confidence bands (\emph{just seen}, \emph{fairly sure}, \emph{hazy}); hearsay from the partner decays twice as fast as first-hand observation. The store holds at most 24 entries, at most 16 from any one author. Nothing in it is validated against the simulator: it records beliefs, and a stale or mistaken row is part of what the condition tests. Both conditions retain the scene image, legend, live-state JSON, operation list, failure feedback, and each player's private six-response action history. \rev{The notebook is the only channel for what the partner has \emph{observed}; the partner's current operation and claims are visible in the live state in both conditions.}

\paragraph{Results.} Shared memory does not help uniformly. \rev{Its effect differs in sign across models and is not statistically reliable (Figure~\ref{fig:ablation}, left). Scores change by $+4.8\pm2.2$ for \texttt{claude-opus-5} (3$\rightarrow$7 orders), $+5.2\pm4.6$ for \texttt{muse-spark-1.3} (19$\rightarrow$23), and $+1.9\pm3.5$ for \texttt{qwen3.8-27b} (9$\rightarrow$10). They change by $-5.7\pm5.7$ for \texttt{gpt-6-astra} (31$\rightarrow$28) and $-5.8\pm4.6$ for \texttt{gemini-3.5-flash-lite} (10$\rightarrow$6). No paired sign test reaches significance; the smallest $p$ is $0.125$, for \texttt{claude-opus-5} with four maps improved, none worse, and 16 tied. The effect is not ordered by model strength: the second-strongest model gains while the strongest loses. For \texttt{gemini-3.5-flash-lite} and \texttt{qwen3.8-27b}, the notebook lowers aim accuracy by $10.1$ and $8.6$ points and adds about 8 failed actions per episode, consistent with note-keeping competing with command selection. The notebook increases input tokens by 14--22\%. On this evidence, the notebook is not a reliable improvement for any of the five models.}

\subsection{\rev{Prompt-Text Ablation}}
\label{sec:textablation}

\begin{revised}
To measure how much performance depends on textual scaffolding, the minimal condition replaces the system prompt of about 3,600 words with one of about 440 words and keeps the images unchanged. The notebook is off in both conditions. It keeps roughly what a human player would have (menu and recipes, the pay rule, a second chef sharing the score, bare operation names, and the grid-label convention) and removes all station descriptions, strategy, and partner information. Failure feedback and the private history remain.

Input tokens fall by 42--50\%. Scores fall for \texttt{gpt-6-astra} ($-14.6\pm6.6$; 31$\rightarrow$21 orders) and \texttt{gemini-3.5-flash-lite} ($-10.3\pm4.3$; 10$\rightarrow$3), whose aim accuracy drops by $11.3$ and $16.5$ points. The other three change by at most two points and $1.3$ points of completion (Figure~\ref{fig:ablation}, right). Neither decline is significant under the sign test ($p=0.21$ and $0.11$). The ablation measures the combined effect of reducing text assistance; the lower aim accuracy does not show how much of either score change comes from command validity. With the minimal prompt, \texttt{gpt-6-astra} ($26.3\%$) and \texttt{muse-spark-1.3} ($23.8\%$) still complete more orders than the privileged script. We did not run an image-removal condition, so the contribution of vision remains unmeasured.
\end{revised}



\section{Conclusion and Future Work}

This paper develops the design of an Overcooked-inspired benchmark for visual planning and \rev{two-player} coordination in agent-generated Unreal Engine kitchens. Its 20 maps span five categories covering independent work, shared equipment, changing terrain, and required item transfers. \rev{On all 20 maps, five VLMs complete between $7.5\%$ and $35.0\%$ of orders. The two strongest exceed a privileged scripted reference, and every map has at least one demonstrated delivery. A shared self-authored notebook does not reliably help any model, and reducing the prompt to minimal text lowers the score of two of five models.}

\begin{revised}
\paragraph{Scope and limitations.} DuoCook evaluates image-grounded target selection and scheduling by one model controlling two players, with engine-executed navigation, paused inference, and structured order and partner state. It does not test low-level control, real-time reaction, communication between independently controlled agents, or perception without textual support. With one run per map and condition, rankings and ablation effects are descriptive. Five models do not support general claims about how memory or prompt text interacts with model capability, and the map categories do not isolate individual mechanisms such as contention or terrain motion. The scripted reference establishes attainability, not optimality, and the reported metrics measure outcomes and command validity rather than coordination quality directly.
\end{revised}







\subsection*{AI Use Statement}

We used \texttt{claude-opus-5} to author scene-generation prompts and construct benchmark maps, as described in Section~\ref{sec:generation}. We also used AI tools for retrieval and discovery of related work and to polish the manuscript.




\bibliography{iclr2027_conference}
\bibliographystyle{iclr2027_conference}

\appendix
\section{Map Correspondence}
\label{app:maps}

\rev{Table~\ref{tab:map_correspondence} lists all 20 functional maps and their categories. Every map is used in every experiment: the main condition, both ablations, and the scripted reference.}

Evaluation identifiers use underscores in place of source-level hyphens, for example \texttt{1\_3}. Festive 1-6 and Festive 1-7 correspond to \texttt{festive\_1\_6} and \texttt{festive\_1\_7}; Kevin 1 corresponds to \texttt{kevin\_level\_one}.

\begin{table}[htbp]
    \centering
    \caption{\rev{Source-level correspondence and map categories. All 20 maps are evaluated in all conditions. Dynamic maps move every 500 steps.}}
    \label{tab:map_correspondence}
    \rev{\begin{tabular}{lll}
        \hline
        Map category & Source levels & Maps \\
        \hline
        Basic & 1-1, 1-3, 1-4, 1-5, Kevin 1 & 5 \\
        Shared equipment & 3-1, 6-4, Festive 1-6, Festive 1-7 & 4 \\
        Dynamic & 1-6, 2-1, 3-3, 6-1 & 4 \\
        Dynamic with shared equipment & 3-4, 4-1, 4-3, 6-3 & 4 \\
        Separated workspaces & 2-3, 2-4, 3-2 & 3 \\
        \hline
        Total & & 20 \\
        \hline
    \end{tabular}}
\end{table}

\section{\rev{Per-Map Results}}
\label{app:permap}

\begin{figure}[htbp]
    \centering
    \includegraphics[width=\linewidth]{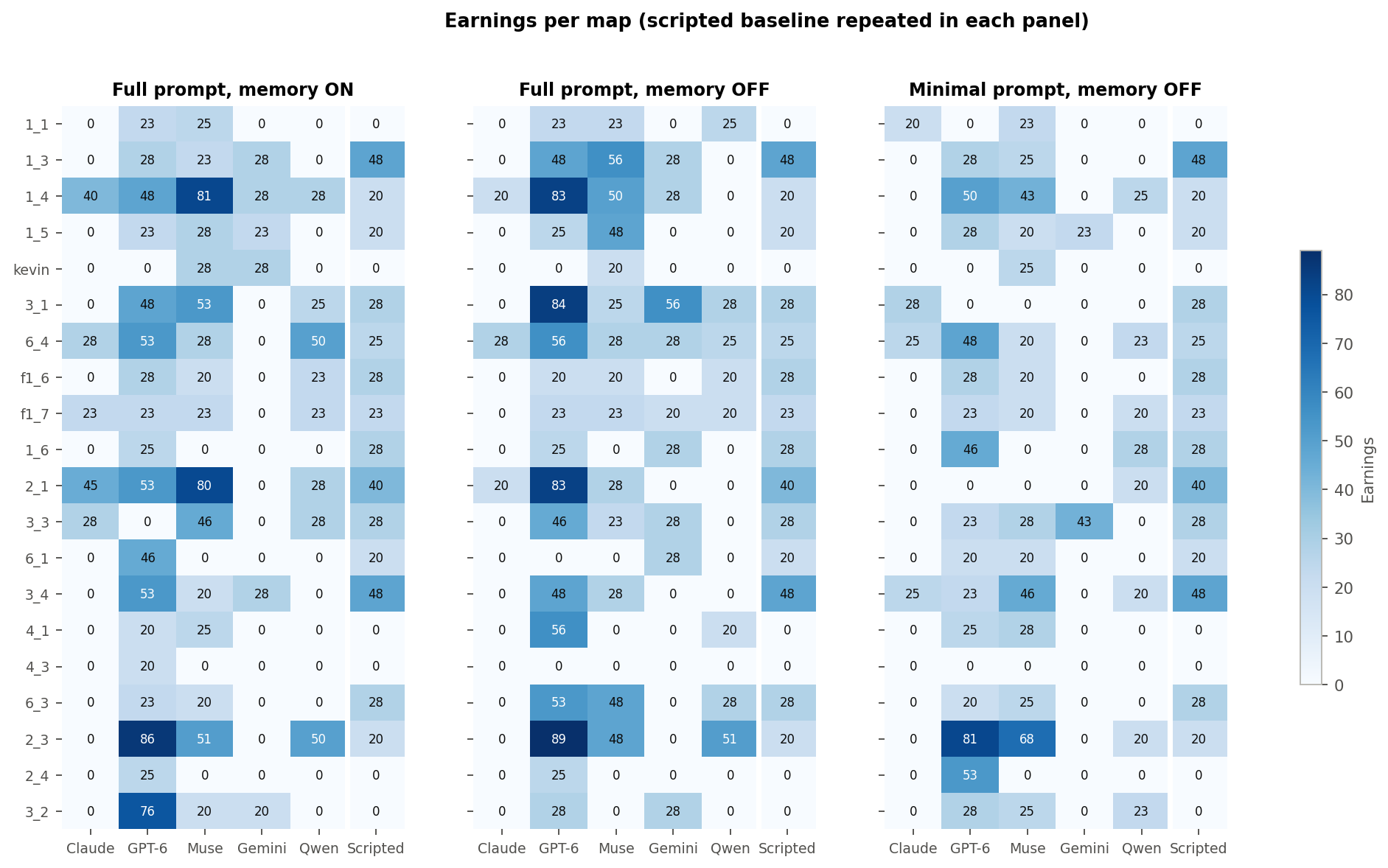}
    \caption{\rev{Score per map, model, and condition. The privileged scripted reference, which has no notebook and no prompt, is repeated in each panel. GPT-6 denotes \texttt{gpt-6-astra}.}}
    \label{fig:heatmap}
\end{figure}

\begin{figure}[htbp]
    \centering
    \includegraphics[width=0.75\linewidth]{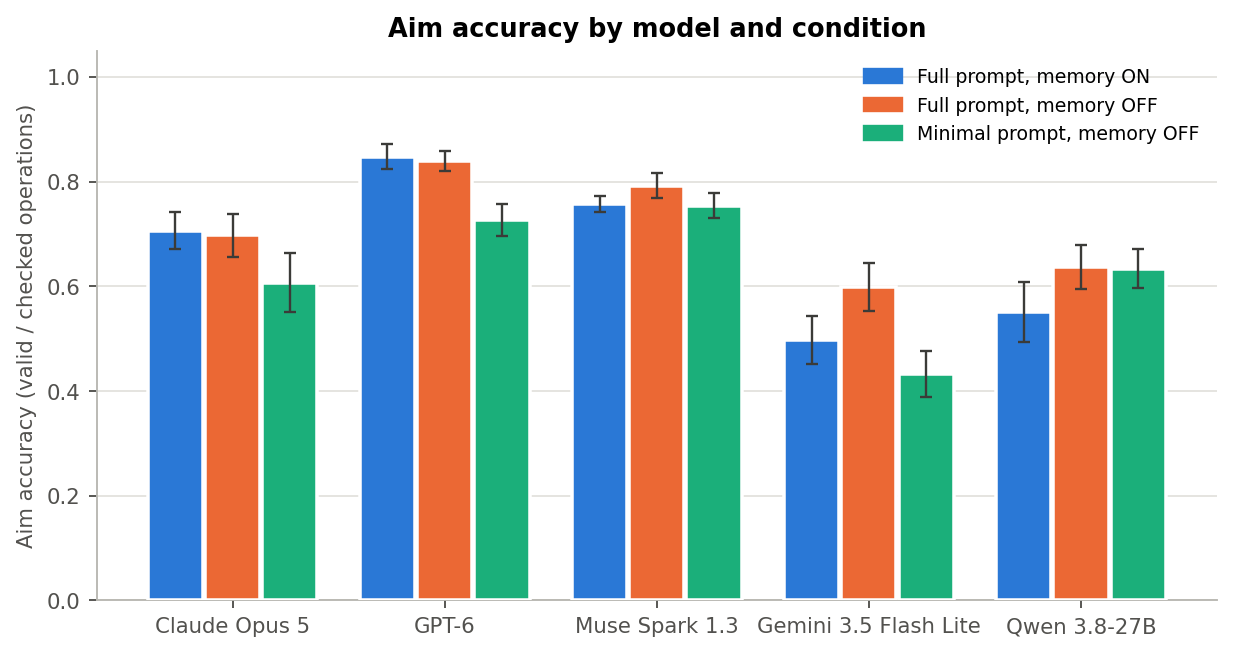}
    \caption{\rev{Aim accuracy (valid / graded commands), mean $\pm$ standard error across the 20 maps. The scripted reference does not aim and is not graded.}}
    \label{fig:aim}
\end{figure}

\begin{revised}
Figure~\ref{fig:heatmap} gives the score of every episode, and Figure~\ref{fig:aim} gives aim accuracy by model and condition. Each episode seats four orders, and a score of 20--28 corresponds to one delivery. All values come from one episode per model, map, and condition. These 320 episodes replace the results of the submitted version, whose source configurations and episode records could not be recovered; the two versions also differ in map set and in how delivery precision is aggregated.
\end{revised}

\section{\rev{Map Construction Records}}
\label{app:generation}

\begin{revised}
Each level begins from a reference image and a written specification derived from it. The specification includes an asset palette verified on disk, an explicit build order (lighting, structure, interactable stations, props, serving, players, cameras, visibility, verification, then decoration), and a numbered verification checklist of 27 to 39 items. The construction agent executes the specification by issuing short Python scripts into the live Unreal Engine editor through an MCP tool, reading back actor bounds, labels, and folders to verify intermediate results. After the rule-based checklist, the level is rendered from top-down and isometric cameras and compared with the reference, and the agent or operator applies repairs. Logs do not distinguish operator fixes applied through the same tool from agent edits.

Table~\ref{tab:generation} reports surviving per-level records for the benchmark levels. No records survive for level 1-1. Costs are cumulative and include discarded builds. A cost or run count of zero means that no model-cost record could be attributed to that level, not that construction was free. The checklist result was written into the export manifest for only two benchmark levels; on five others, the surviving log records the checklist with no failed item.

Recurring failures, now encoded in the specifications, include unlit levels that render black without an error, invented asset paths that spawn invisible actors, editor crashes from loading levels through the tool, silent save failures on existing maps, spawns landing in a sublevel, rotator argument-order errors, and concurrent edits in a shared editor.
\end{revised}

\begin{table}[htbp]
    \centering
    \small
    \caption{\rev{Construction records per benchmark level. Subs.: prompt submissions; Runs: agent runs; Calls: Python script calls; Err.: tool errors; Ep.: build episodes, meaning returns to a level after more than one hour; Check: verification checklist result where recorded.}}
    \label{tab:generation}
    \rev{\begin{tabular}{lrrrrrrl}
        \toprule
        Level & Subs. & Runs & Calls & Err. & Ep. & Cost (\$) & Check \\
        \midrule
        1-3 & 9 & 5 & 75 & 3 & 1 & 8.31 & --- \\
        1-4 & 16 & 8 & 115 & 4 & 4 & 10.85 & --- \\
        1-5 & 1 & 8 & 82 & 3 & 2 & 8.72 & --- \\
        Kevin 1 & 5 & 7 & 83 & 2 & 2 & 11.92 & --- \\
        3-1 & 2 & 0 & 36 & 0 & 1 & 0.00 & --- \\
        6-4 & 17 & 6 & 72 & 2 & 4 & 14.02 & 32/32 \\
        Festive 1-6 & 1 & 1 & 110 & 0 & 1 & 19.10 & --- \\
        Festive 1-7 & 2 & 0 & 132 & 1 & 1 & 0.00 & --- \\
        1-6 & 25 & 10 & 109 & 3 & 2 & 13.13 & --- \\
        2-1 & 9 & 2 & 128 & 1 & 3 & 13.14 & --- \\
        3-3 & 6 & 5 & 75 & 1 & 1 & 16.28 & 0 failed \\
        6-1 & 3 & 1 & 66 & 3 & 1 & 10.55 & 0 failed \\
        3-4 & 8 & 5 & 74 & 1 & 1 & 11.65 & 0 failed \\
        4-1 & 3 & 2 & 55 & 2 & 1 & 8.03 & 0 failed \\
        4-3 & 7 & 3 & 81 & 5 & 2 & 12.06 & 0 failed \\
        6-3 & 6 & 5 & 127 & 4 & 1 & 20.05 & 32/32 \\
        2-3 & 8 & 3 & 121 & 6 & 1 & 10.91 & --- \\
        2-4 & 4 & 0 & 47 & 0 & 1 & 0.00 & --- \\
        3-2 & 18 & 7 & 163 & 45 & 3 & 11.84 & --- \\
        \bottomrule
    \end{tabular}}
\end{table}

\section{\rev{API Cost Estimation}}
\label{app:cost}

\begin{revised}
The API gateway used for four of the five models does not report billing. For these models, we estimate the cost of each episode from the recorded uncached input, cached input, and output tokens (including reasoning tokens), summed over all calls including retries, at the providers' list prices (Table~\ref{tab:prices}). For \texttt{qwen3.8-27b}, accessed through OpenRouter, we use the amount actually billed, read from the account balance before and after each episode; list-price arithmetic underestimates this bill because OpenRouter may route requests to more expensive providers. The scripted reference makes no API calls.
\end{revised}

\begin{table}[htbp]
    \centering
    \small
    \caption{\rev{List prices (USD per million tokens) and mean API cost per episode (USD) over the 20 maps in each condition. $^\dagger$Billed by OpenRouter.}}
    \label{tab:prices}
    \rev{\begin{tabular}{lrrrrrr}
        \toprule
        & \multicolumn{3}{c}{Price per million tokens} & \multicolumn{3}{c}{Cost per episode} \\
        \cmidrule(lr){2-4}\cmidrule(lr){5-7}
        Model & Input & Cached & Output & Notebook on & Notebook off & Minimal \\
        \midrule
        \texttt{gpt-6-astra} & 10.00 & 1.00 & 50.00 & 5.41 & 5.01 & 3.32 \\
        \texttt{claude-opus-5} & 5.00 & 0.50 & 25.00 & 4.61 & 4.06 & 1.93 \\
        \texttt{muse-spark-1.3} & 1.25 & 0.15 & 4.25 & 1.09 & 1.12 & 0.86 \\
        \texttt{gemini-3.5-flash-lite} & 0.30 & 0.03 & 2.50 & 0.33 & 0.29 & 0.17 \\
        \texttt{qwen3.8-27b} & \multicolumn{3}{c}{billed$^\dagger$} & 0.30 & 0.28 & 0.24 \\
        \bottomrule
    \end{tabular}}
\end{table}

\section{\rev{Comparison with Related Cooking Environments}}
\label{app:comparison}

\begin{revised}
Table~\ref{tab:comparison} compares DuoCook with the closest cooking environments along the dimensions that determine what an evaluation measures. Entries for other environments summarize the settings reported in their papers. No single dimension is unique to DuoCook. What it adds is their combination within one interface: rendered observations from which targets are localized mainly visually, terrain that changes during an episode, equipment contention, and mandatory handoffs, each represented across the suite.
\end{revised}

\begin{table}[htbp]
    \centering
    \scriptsize
    \caption{\rev{Comparison with related cooking environments, as reported in each paper. Contention: agents compete for shared stations. Handoff: some tasks cannot be completed without passing items between agents. Observation includes what each evaluated agent sees and which state is private. For DuoCook, counts give the number of maps with each property. ---: not applicable, since only one agent is evaluated.}}
    \label{tab:comparison}
    \setlength{\tabcolsep}{3pt}
    \newcolumntype{L}[1]{>{\raggedright\arraybackslash}p{#1\linewidth}}
    \rev{\begin{tabular}{L{0.12}L{0.12}L{0.14}L{0.13}L{0.08}L{0.08}L{0.07}L{0.13}}
        \toprule
        Environment & Observation & Agents and control & Action level & Changing terrain & Conten\-tion & Hand\-off & Main metrics \\
        \midrule
        Overcooked-AI \citep{carroll2019utility} & Symbolic grid state; fully observable & 2, one policy per agent & Grid moves and interact & No & Yes & One layout & Reward \\
        CuisineWorld \citep{gong2024mindagent} & Text state; fully observable & 2+, central LLM dispatcher & High-level (goto, get, put) & No & Yes & No & Collaboration score \\
        Collab-Overcooked \citep{sun2025collab} & Text; shared observation, recipe known to one agent & 2, one LLM per agent, with messages & High-level primitives & No & No (isolated) & Yes, all tasks & Success, process efficiency \\
        Robotouille \citep{gonzalez2025robotouille} & Text predicates & 1 evaluated; multi-agent supported & Grounded high-level & No & --- & No & Success \\
        ParaCook \citep{zhang2026paracook} & Symbolic grid state; full initial state to one planner & 2+, central LLM planner & High-level (MoveTo) & No & Yes & No & Success, completion time \\
        CookBench \citep{cai2025cookbench} & RGB from the agent's camera (Unity) & 1 & Fine-grained; human-assisted navigation and camera in the reported study & No & --- & No & Intent accuracy; dish score (1--5); completion time \\
        \midrule
        DuoCook (ours) & Full top-down image (Unreal) + text state; action history private per player & 2, one VLM, per-player calls & Operation + image-labeled cell & Yes (8) & Yes (8) & Yes (3) & Score, completion, precision, aim accuracy, cost \\
        \bottomrule
    \end{tabular}}
\end{table}



\section{Sample Scene Generation Prompts}
\label{app:scenegenprompt}

\begin{tcolorbox}[
    enhanced,
    breakable,
    colback=gray!8!white,
    colframe=gray!50!black,
    colbacktitle=gray!30!black,
    coltitle=white,
    fonttitle=\bfseries,
    title=Overcooked Level 2-1 Prompt,
    rounded corners,
    boxrule=1pt,
    arc=8pt,
    left=10pt, right=10pt, top=10pt, bottom=10pt
]

\begin{minipage}[t]{0.48\linewidth}
    \centering
    \includegraphics[width=\linewidth, height=0.6\linewidth]{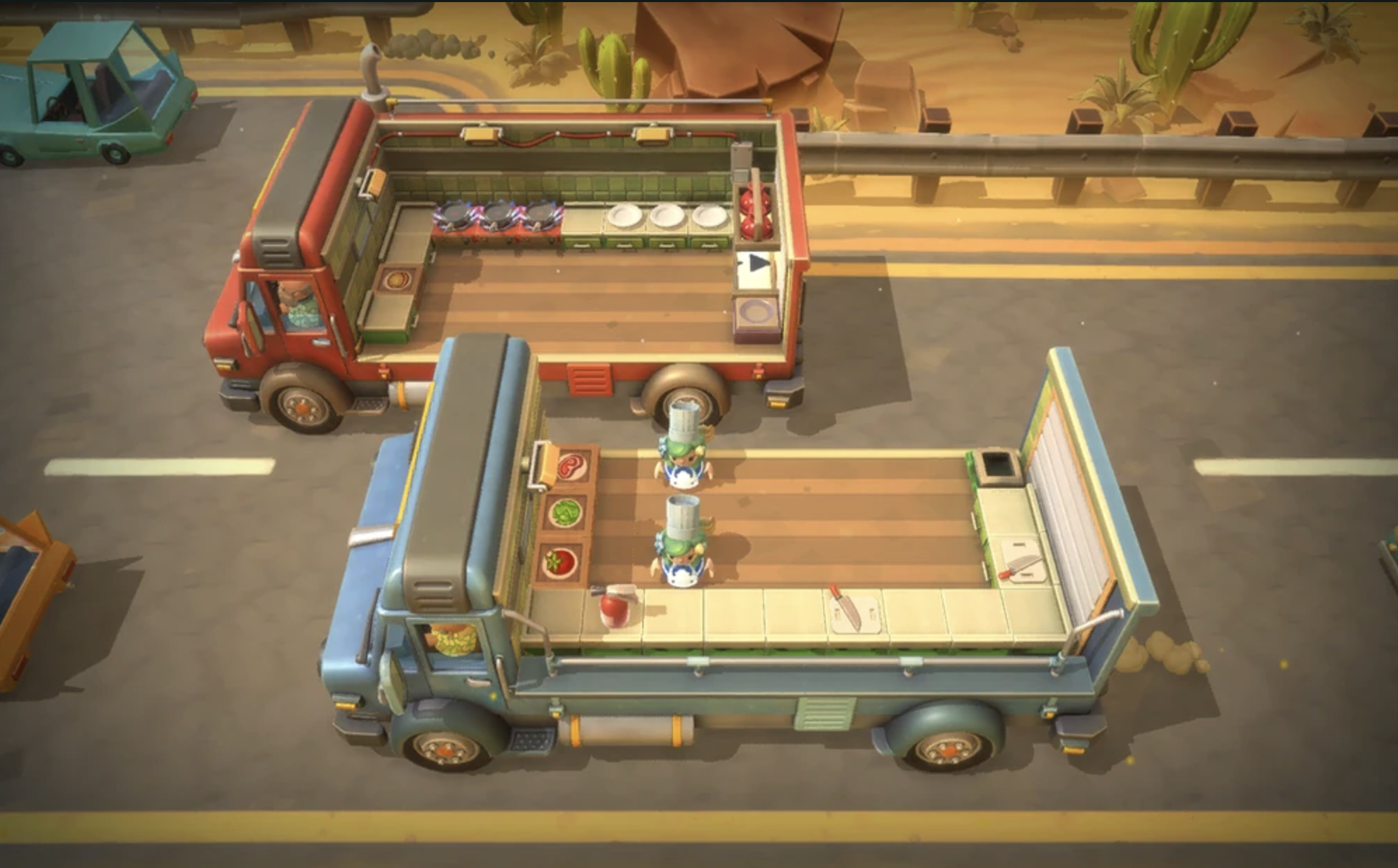}\\[2pt]
    \small\textit{Overcooked Level 2-1}
\end{minipage}
\hfill
\begin{minipage}[t]{0.48\linewidth}
    \centering
    \includegraphics[width=\linewidth, height=0.6\linewidth]{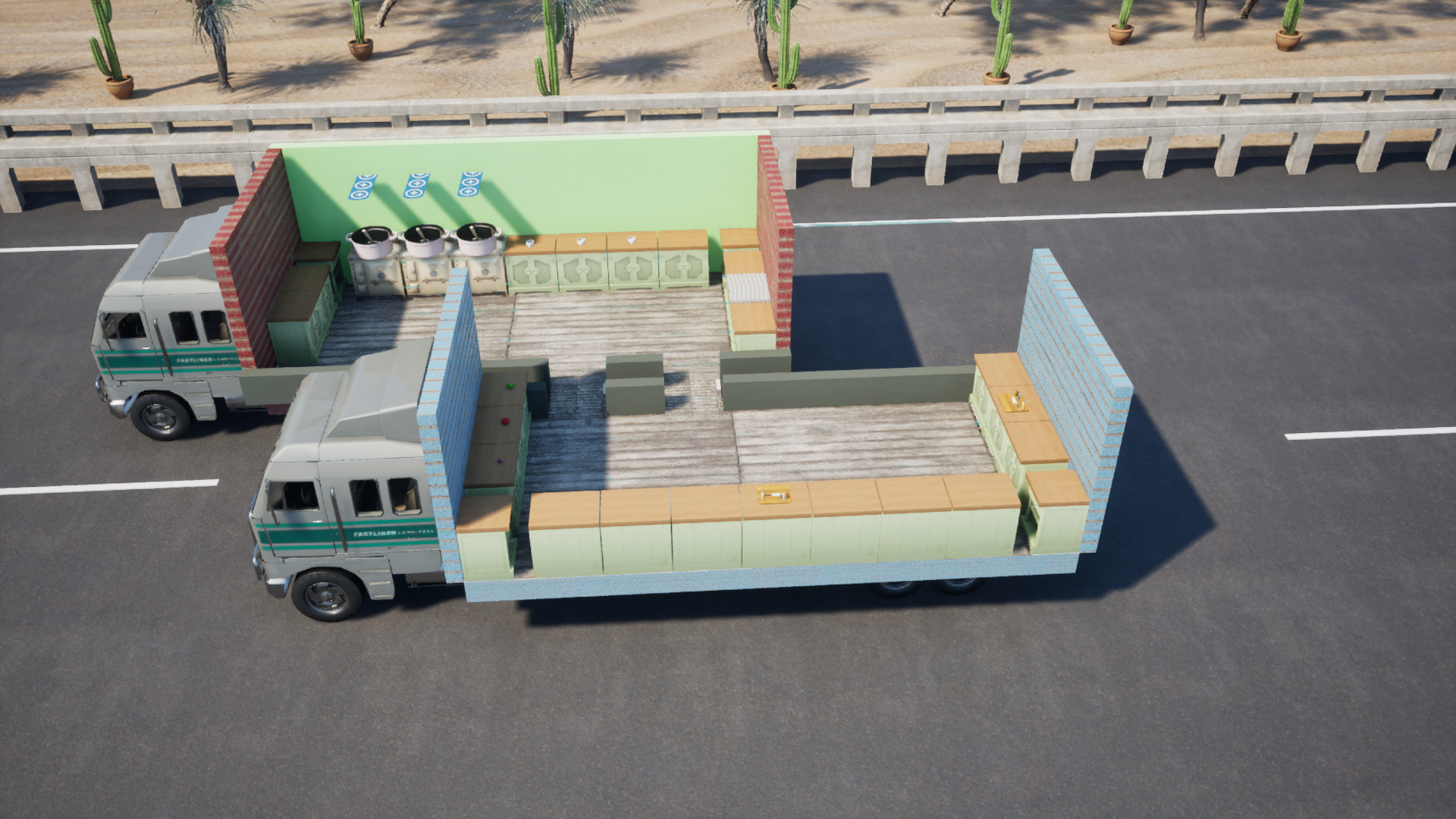}\\[2pt]
    \small\textit{Generated Image}
\end{minipage}

\vspace{10pt}
\hrule
\vspace{8pt}

\textbf{Prompt:}

\begin{lstlisting}
Overcooked level 2-1 -- two flatbed trucks on a desert highway

Target: /Game/SavedScenes/overcooked_level_2_1
Reference: overcooked_level_2_1.png (Overcooked 2, level 2-1)

GOAL: A level with ~149 actors where two flatbed truck beds sit side by
side on a desert highway, staggered along the road, with open asphalt
between them crossed by exactly two gangplanks.

- North (red) truck cooks and serves: 3 stoves, 3 clean bowls, the
  delivery belt, and the plate return.
- South (blue) truck preps: 3 ingredient stations, 2 cutting boards,
  and a long run of free counters.
- Both trucks use a real cab mesh (SM_vehTruck_vehicle08_No_Wheel)
  with 6 real wheel meshes each, positioned from the skeletal mesh's
  sockets. Everything else (deck, skirt, walls, rails, gangplanks) is
  built from Cube/Plane primitives with materials applied.
- The desert reads as a desert: asphalt, lane paint, guardrail, cacti,
  mesas.
- No signage of any kind -- a station is its actor label and nothing
  else.

KEY DESIGN DECISIONS:
1. Decks sit at z = 0; the road is 150 units below (z = -150), so all
   standard kitchen seating heights (counter 100, hob 100, bowl
   104.17, etc.) apply unchanged.
2. The gap between the trucks is real open road -- not a trench --
   crossed only by two gangplanks. Do not widen the gap or add a
   third crossing; the bottleneck is the point of the level.
3. All three ingredients are on the blue truck and all three stoves
   are on the red truck, so every dish crosses a gangplank at least
   twice.

LAYOUT: Built on a 13-row x 9-column grid (BP_Grid, tile size 102cm).
Red truck occupies columns 1-4, rows 1-9. Blue truck occupies columns
6-9, rows 5-13. Column 5 is the open gap. Counters, stoves, cutting
boards, and ingredient stations are placed at specific grid cells
following the standard kitchen palette (/Game/KitchenAssets/,
/Game/kitchenDatabase/Interactable/) -- never substituted with
look-alike assets from elsewhere in the project.

CAMERA: top_down camera at (600, 561, 1500), pitch -90, yaw -90,
28mm focal length -- offset from the grid's true center so the red
truck's cab (which extends west of the grid) isn't cut off frame.

DELIVERABLE: The finished level, verified against a measured
checklist (actor counts, seating heights, tag/label correctness,
hidden-at-open state props), plus a top-down screenshot compared
against the reference image.
\end{lstlisting}

\end{tcolorbox}

\begin{tcolorbox}[
    enhanced,
    breakable,
    colback=gray!8!white,
    colframe=gray!50!black,
    colbacktitle=gray!30!black,
    coltitle=white,
    fonttitle=\bfseries,
    title=Overcooked Level 6-4 Prompt,
    rounded corners,
    boxrule=1pt,
    arc=8pt,
    left=10pt, right=10pt, top=10pt, bottom=10pt
]

\begin{minipage}[t]{0.48\linewidth}
    \centering
    \includegraphics[width=\linewidth, height=0.6\linewidth]{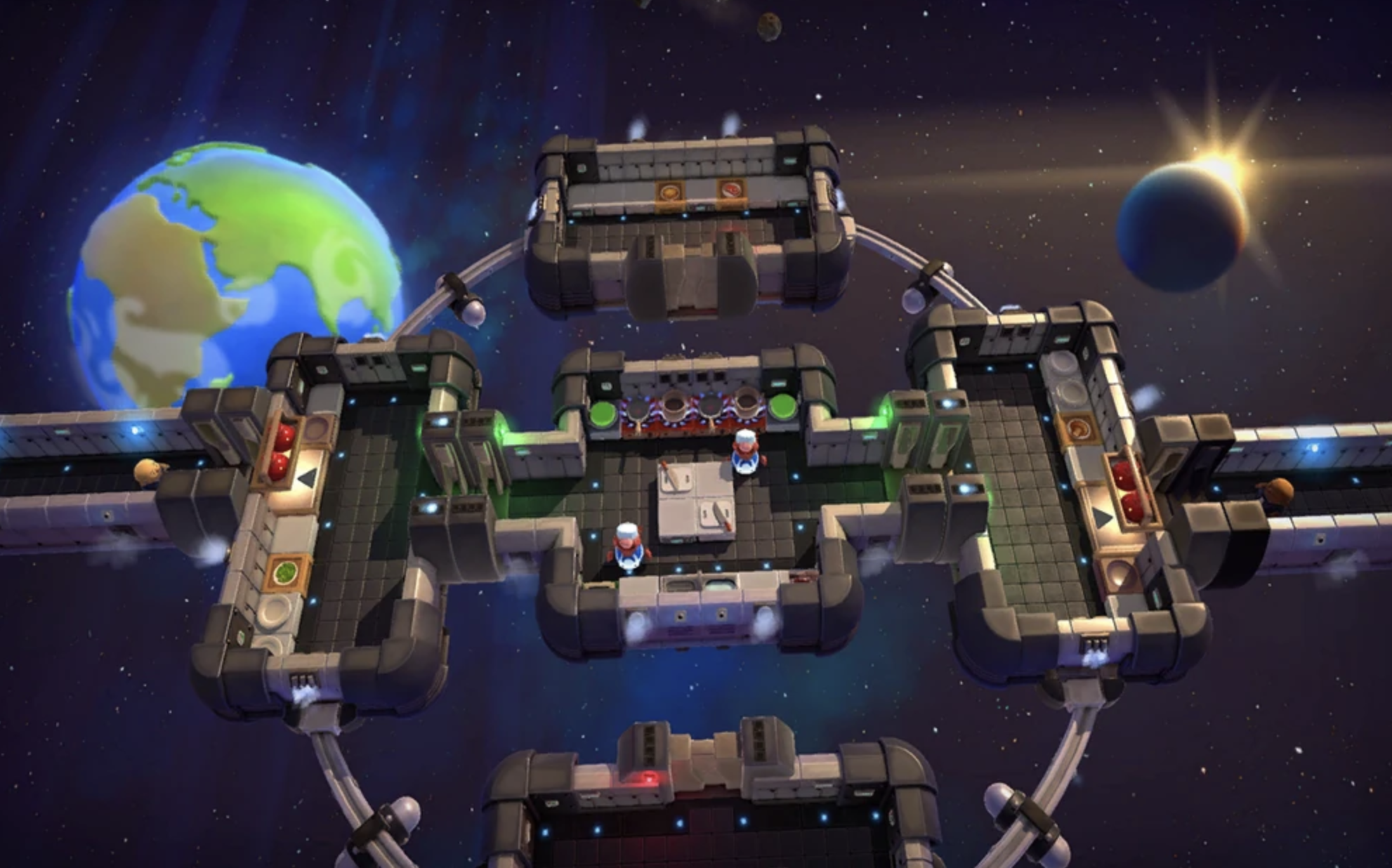}\\[2pt]
    \small\textit{Overcooked Level 6-4}
\end{minipage}
\hfill
\begin{minipage}[t]{0.48\linewidth}
    \centering
    \includegraphics[width=\linewidth, height=0.6\linewidth]{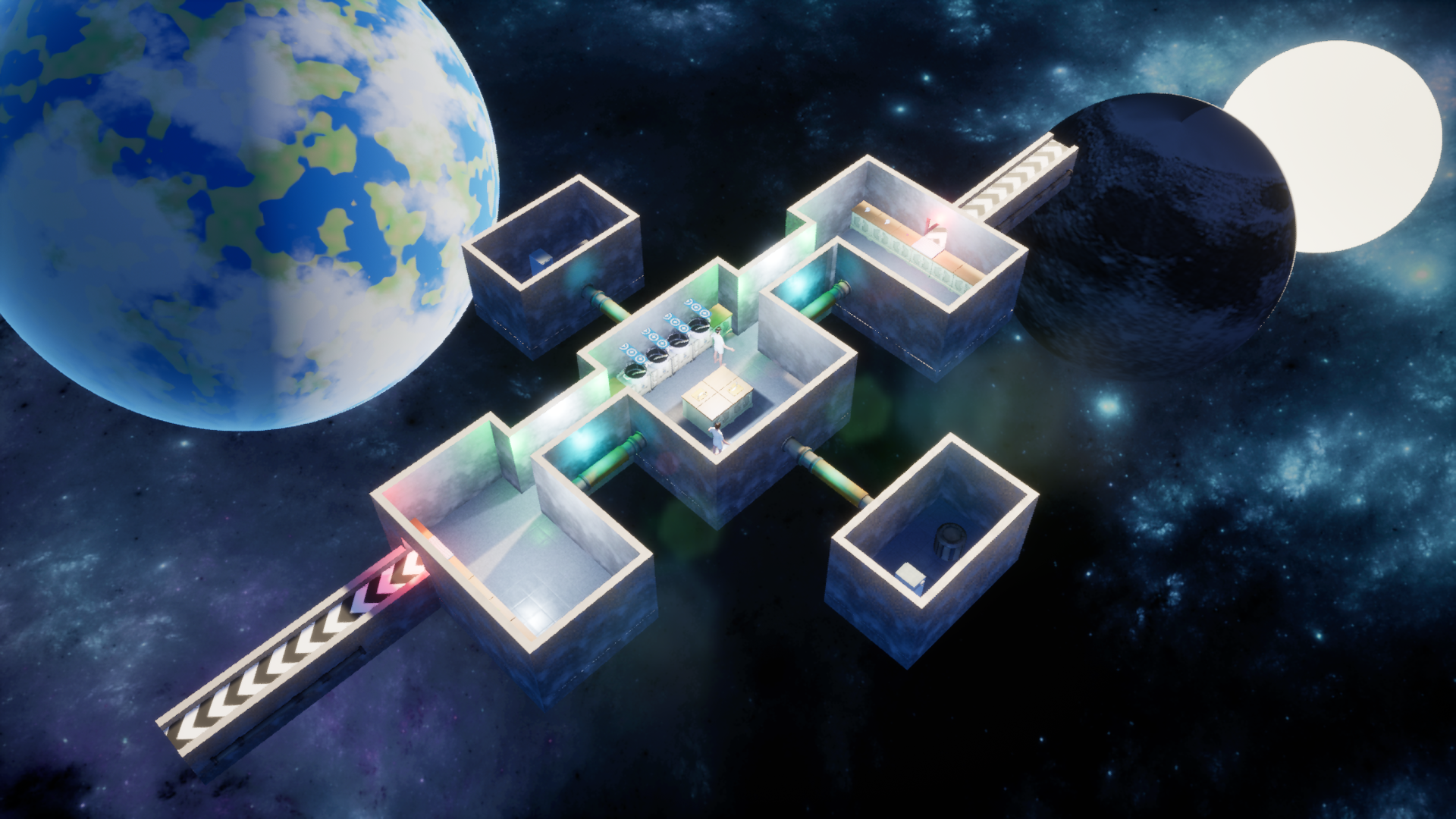}\\[2pt]
    \small\textit{Generated Image}
\end{minipage}

\vspace{10pt}
\hrule
\vspace{8pt}

\textbf{Prompt:}

\begin{lstlisting}
Overcooked level 6-4 -- "Space Station: the Docked Ring" -- scene prompt

Target: /Game/SavedScenes/overcooked_level_6_4
Reference: overcooked_level_6_4.png (Overcooked! 2 level 6-4)

GOAL: A kitchen space station with ~52 actors: a central hub for
cooking, two side modules (west and east) docked to it via short
corridors for serving and prep, and two unreachable decorative ring
pods (north and south) that never turn in this build -- the station
stays static.

STATIONS:
- 4 stoves in the hub's north run, each with a full pot/spatula/pot
  counter stack (stove_1..4, tags Stove_1..4).
- 2 cutting boards + knives on a 2x2 island in the hub's center.
- 3 ingredient stations (lettuce west, onion + tomato east), each
  with a pickup, 2 whole and 2 sliced ingredient props.
- 1 delivery belt + pusher in the west module (dishes exit west);
  the east "hatch" is decorative -- just plain counters.
- 1 sink in the hub's south run; 1 plate return in the west module.
- 4 clean bowls (2 west, 2 east).

LAYOUT: Built on an 18-column x 7-row grid (BP_Grid, 102cm tiles).
West module (columns 1-4) and east module (columns 15-18) dock to a
central hub (columns 7-12) via corridors (columns 5-6 and 13-14).
Two ring pods sit off-grid to the north and south, closed and
decorative only.

KEY DESIGN DECISIONS:
1. The station doesn't rotate -- both side modules stay permanently
   docked, matching the reference's single frame.
2. All stovetop items (pots and pans in the reference) become
   identical stove stations for consistency.
3. Only one delivery belt and one plate return are built (west
   module); the reference's second hatch/return on the east side
   are represented as plain, non-functional counters.
4. Ingredients: lettuce (west crate), onion (east crate), and tomato
   (added to an east counter, since the reference's north-pod crates
   are unreachable).
5. Decoration (skybox, passing ship, pod crates/consoles, hull
   pipes) is added last, from existing library assets only -- never
   built from primitives, and always with collision disabled.

STRUCTURE: Floors, walls, and metal hull plating define three rooms
and two corridors, all built from primitives (the only place
primitives are allowed). Five invisible collision blockers seal off
non-walkable empty cells so the level doesn't leak walkable space
outside the intended rooms.

LIGHTING: No sky atmosphere or fog (it's space) -- lighting comes
from a directional "sun," a skylight, and five point lights placed
through the station's rooms for legibility.

CAMERAS: A top-down camera and an isometric camera, both framing the
full station including both ring pods, placed in a dedicated
"Cameras" outliner folder.

DELIVERABLE: The finished level with every station present and
correctly labeled/tagged, props resting properly on their surfaces,
state props (raw/sliced ingredients, soup-in-pot props, dirty bowl)
hidden at level open, verified against a measured checklist, plus
top-down and isometric screenshots compared to the reference image.
\end{lstlisting}

\end{tcolorbox}

\begin{tcolorbox}[
    enhanced,
    breakable,
    colback=gray!8!white,
    colframe=gray!50!black,
    colbacktitle=gray!30!black,
    coltitle=white,
    fonttitle=\bfseries,
    title=Overcooked Festive Level 1-7 Prompt,
    rounded corners,
    boxrule=1pt,
    arc=8pt,
    left=10pt, right=10pt, top=10pt, bottom=10pt
]

\begin{minipage}[t]{0.48\linewidth}
    \centering
    \includegraphics[width=\linewidth, height=0.6\linewidth]{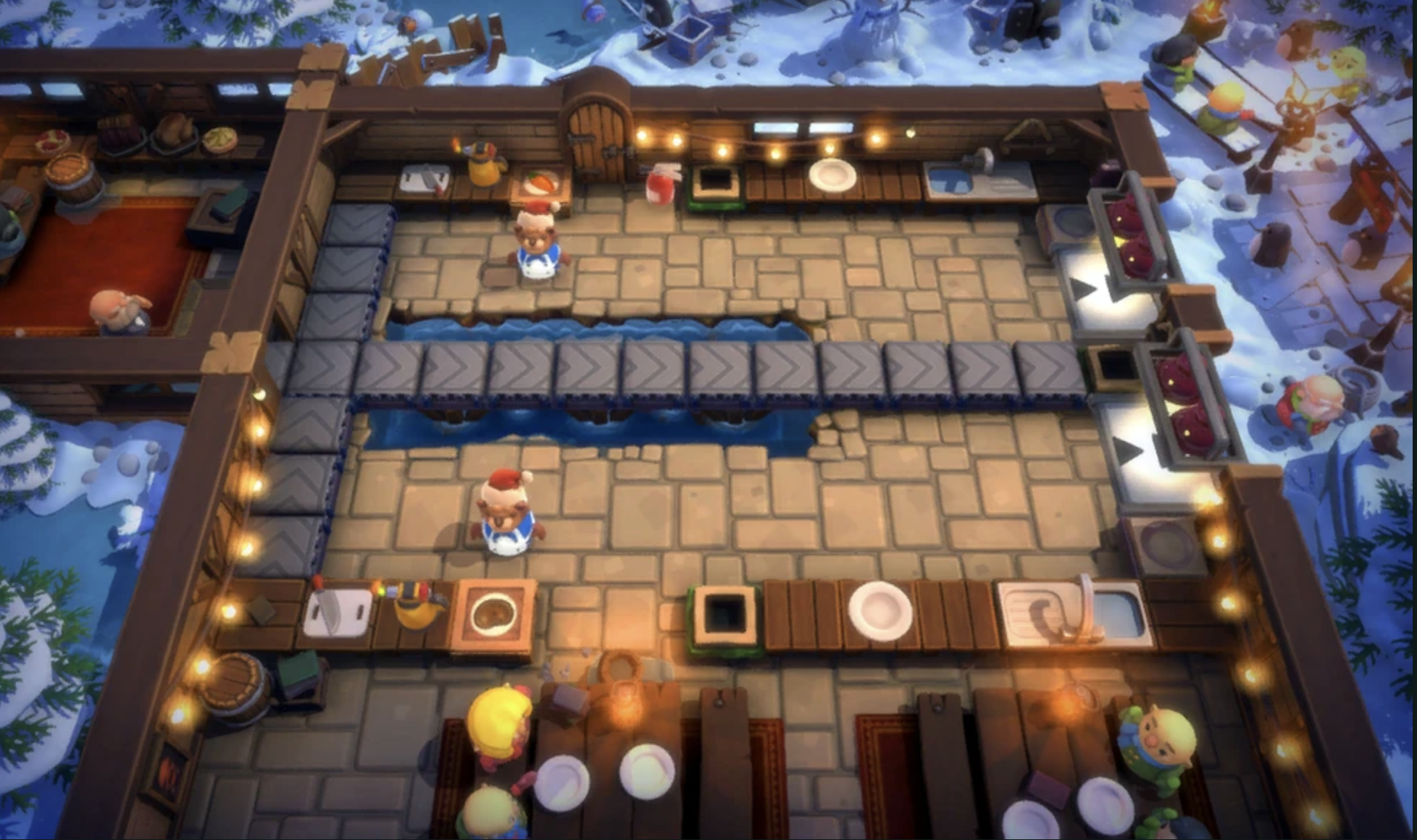}\\[2pt]
    \small\textit{Overcooked Festive Level 1-7}
\end{minipage}
\hfill
\begin{minipage}[t]{0.48\linewidth}
    \centering
    \includegraphics[width=\linewidth, height=0.6\linewidth]{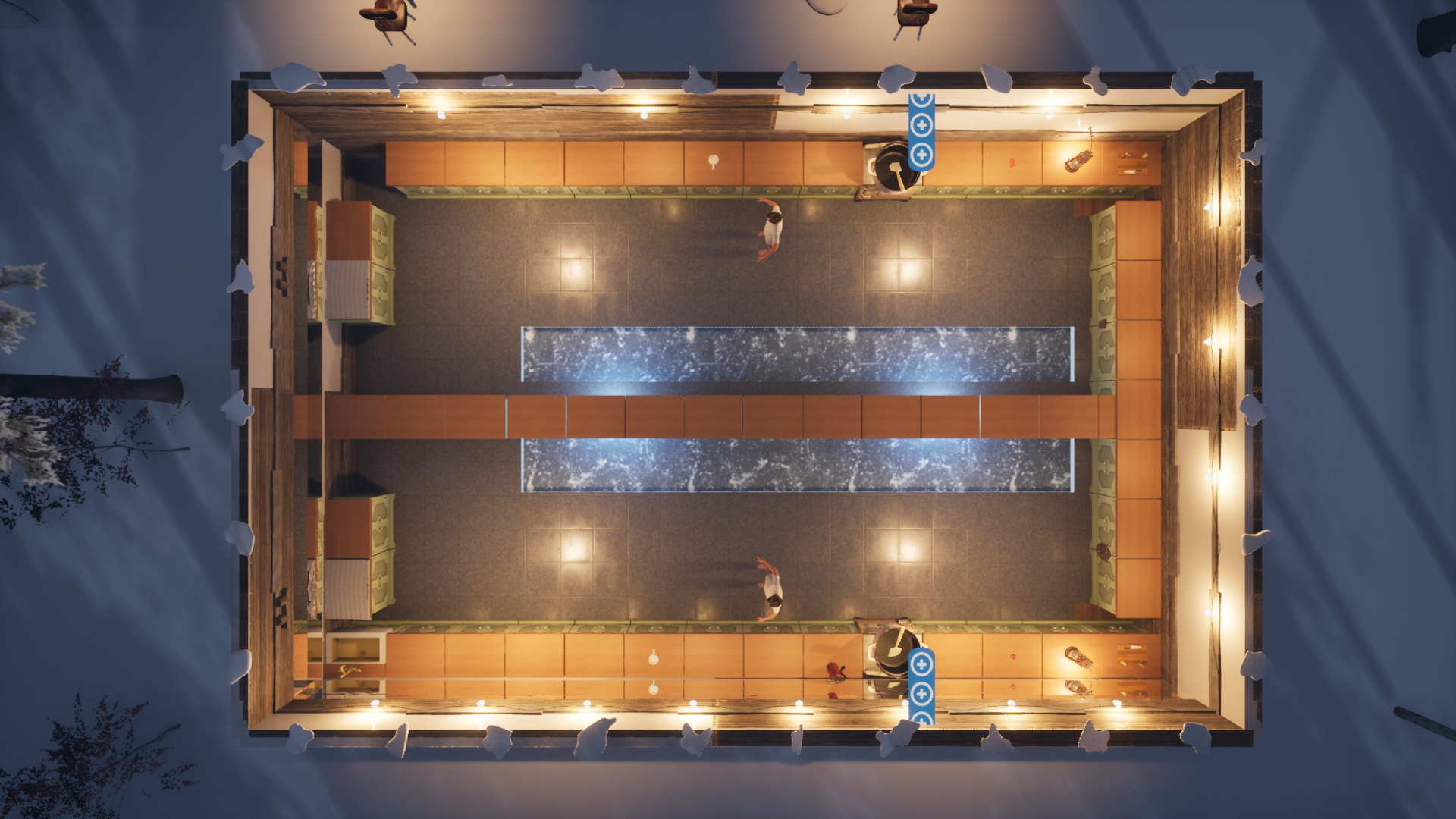}\\[2pt]
    \small\textit{Generated Image}
\end{minipage}

\vspace{10pt}
\hrule
\vspace{8pt}

\textbf{Prompt:}

\begin{lstlisting}
Overcooked festive 1-7 -- "The Frozen Brook Lodge" -- scene prompt

Target: /Game/SavedScenes/overcooked_festive_level_1_7
Reference: overcooked_level_festive_1_7.png (Overcooked! festive level)

GOAL: A ~90-actor timber lodge kitchen split in half by a frozen
brook running east-west across the floor. A chevron-block bridge
crosses the brook mid-room, and a second walkway runs up the west
wall. Each half of the kitchen is a mirror of the other, connected
by the bridge and a walkable strip at the east end of the room.

STATIONS:
- 2 stoves, one per half, each with a full pot/spatula/pot counter
  stack (stove_1, stove_2).
- 2 cutting boards + knives, one per half.
- 2 ingredient stations: onion (north half, from the reference's
  carrot crate) and meat (south half).
- 2 separate delivery belts + pushers on the east wall, one per
  half -- dishes exit east through two chutes.
- 1 sink (north-east corner) and 1 plate return (east wall, north
  half) -- sinkless-style return using a single hidden dirty-bowl
  prop.
- 2 clean bowls, one per half.

LAYOUT: A 14-column x 9-row grid (BP_Grid, 102cm tiles). North and
south counter runs face each other across the room; the brook cuts
across the middle two rows, blocked to movement except at the
bridge and the open strip at the east wall.

KEY DESIGN DECISIONS:
1. Two stoves rather than the reference's single hob, so each half
   of the split kitchen has its own -- one per side of the water.
2. Two ingredients: onion in the north half, meat in the south
   half, each with its own cutting board.
3. The reference's "bin" objects become plain, non-functional
   counters; kettles, a fire extinguisher, and bauble racks are
   decoration only.
4. The brook is built from water planes with an invisible
   collision blocker -- chefs cross only via the bridge or the
   east-end strip, never through the water.
5. No off-the-shelf Christmas assets exist in the library (no
   snowman, wreath, or bauble meshes) -- the festive read comes
   from warm string-light lighting against blue snow outside,
   not literal ornaments.

STRUCTURE: A stone floor (deliberately cool-toned, not warm, so it
reads distinctly from the timber counters and walls), four timber
walls with a windowed band, and a snow-covered exterior are all
built from primitives -- the only place primitives are allowed.
Everything else is library assets.

LIGHTING: Warm string-light and hearth-style point lights inside,
a dim cool directional "sun" plus fog for a blue-hour winter
atmosphere, and a small cold light glow along the brook itself.

CAMERAS: A top-down and an isometric camera, both in a dedicated
"Cameras" folder, framing the full room including the brook.

DELIVERABLE: The finished level with both cooking/prep/serving
setups complete and correctly labeled, all props resting properly
on their surfaces, state props (sliced ingredients, cooked soup
overlays, the dirty bowl) hidden at level open, both halves of the
kitchen reachable from both chef spawns, checked against a measured
checklist, plus top-down and isometric screenshots compared to the
reference image.
\end{lstlisting}

\end{tcolorbox}

\end{document}